\documentclass{article}
\usepackage{spconf,amsmath,amssymb,graphicx}
\usepackage{adjustbox}
\usepackage[table,xcdraw]{xcolor}
\usepackage{booktabs}
\usepackage{url}
\usepackage[table]{xcolor}

\title{Investigating Training Strategies and Model Robustness of Low-Rank Adaptation for Language Modeling in Speech Recognition}
\name{ \begin{tabular}{@{}c@{}}
Yu Yu$^{1^{*},2}$\thanks{$^*$Work done mainly while the first author was an intern at Amazon.}, C.-H. Huck Yang$^{1}$,
Tuan Dinh$^{1}$, Sungho Ryu$^{1}$, Jari Kolehmainen$^{1}$, Roger Ren$^{1}$, Denis Filimonov$^{1}$ \\
Prashanth G. Shivakumar$^{1}$, Ankur Gandhe$^{1}$,  Ariya Rastrow$^{1}$, Jia Xu$^{2}$, Ivan Bulyko$^{1}$, Andreas Stolcke$^{1}$
\end{tabular}}
\address{$^{1}$Amazon Alexa AI, USA\\$^{2}$Stevens Institute of Technology, USA}
\begin{document}
\ninept
\maketitle
\begin{abstract}
The use of low-rank adaptation (LoRA) with frozen pretrained language models (PLMs) has become increasing popular as a mainstream, resource-efficient modeling approach for memory-constrained hardware. In this study, we first explore how to enhance model performance by introducing various LoRA training strategies, achieving relative word error rate reductions of 3.50\% on the public Librispeech dataset and of 3.67\% on an internal dataset in the messaging domain. To further characterize the stability of LoRA-based second-pass speech recognition models, we examine robustness against input perturbations. These perturbations are rooted in homophone replacements and a novel metric called N-best Perturbation-based Rescoring Robustness (NPRR), both designed to measure the relative degradation in the performance of rescoring models. Our experimental results indicate that while advanced variants of LoRA, such as dynamic rank-allocated LoRA, lead to performance degradation in $1$-best perturbation, they alleviate the degradation in $N$-best perturbation. This finding is in comparison to fully-tuned models and vanilla LoRA tuning baselines, suggesting that a comprehensive selection is needed when using LoRA-based adaptation for compute-cost savings and robust language modeling.
\end{abstract}
\begin{keywords}
Low-rank adaptation, memory-efficient learning, language model rescoring, robust speech recognition.
\end{keywords}
\section{Introduction}
\label{sec:intro}

Automatic speech recognition (ASR) systems~\cite{juang2005automatic} traditionally rely on a combination of acoustic models and language models to convert spoken language to text. While acoustic models focus on the mapping between audio features and phonetic units, language models capture the probabilistic structure of word sequences. One widely adopted approach to enhance recognition accuracy is language model rescoring, where initial transcriptions generated by the ASR system are re-evaluated using a more powerful language model. 

Our study involves the application of a range of low-rank adaptation techniques~\cite{hu2021lora} to enhance the performance of BERT~\cite{devlin2019bert} in rescoring scenarios. These strategies encompass vanilla LoRA, dynamic rank allocation, high-rank warm-up, and mixed-rank training. Subsequently, we assess these models with regard to both performance and robustness. The performance evaluation includes a comprehensive examination of the contribution of individual layers.  To evaluate the resilience of these fine-tuned models against adversarial perturbations, we introduce a novel N-best perturbation algorithm centered around homophone replacement, along with two perturbation strategies termed ``perturb-$1$", perturbing only the hypothesis with the lowest score assigned by the acoustic model, and ``perturb-$N$", perturbing all hypotheses. Additionally, we introduce the concept of $N$-best Perturbation based Rescoring Robustness (NPRR) as a metric to quantify the relative decline in the performance of rescoring models.

Our experimental results suggest that all low-rank adapted language models tend to exhibit diminished robustness compared to fully fine-tuned models. Notably, the training strategy aimed at enhancing performance, specifically dynamic rank allocation, is shown to exacerbate the degradation of adversarial robustness.

Our contributions can be summarized as follows:
\begin{enumerate}
\item We systematically evaluate the performance of the state-of-the-art low-rank adaptation and its advanced variants in rescoring BERT for speech recognition.

\item We conduct a first study to explore the influence of low-rank adaptation training methods on a rescoring model's adversarial robustness. We propose input perturbations based on phonetic similarity and a robustness evaluation metric termed N-best Perturbation based Rescoring Robustness (NPRR). 

\item Our perturbation algorithms probe the stability of low-rank-adapted reranking models and provide insights for future work on robust ASR modeling of $N$-best input.

\end{enumerate}

\section{Related work}
\subsection{ASR Language Modeling Rescoring}
With the recent advances of deep  neural network-based language models, recurrent neural networks~\cite{yang2021multi} (RNNs) and transformers have emerged as effective tools, often outperforming conventional n-gram models in rescoring scenarios~\cite{mikolov2010recurrent}. Specifically, models like BERT and GPT, originating from the natural language processing domain, have been adapted for ASR rescoring, showing significant improvements on various benchmarks~\cite{devlin2019bert}. These neural approaches capture longer contextual information, offering a richer linguistic understanding that can improve the recognition results. For example, RescoreBERT~\cite{xu2022rescorebert} is a popular language model (LM) architecture that uses discriminative training objectives over transformer architectures. However, when the LMs scale up over 100 million trainable parameters, how to efficiently adapt these LMs for ASR tasks has required parameter-efficient adaptation~\cite{gu2023scaling}. 

\subsection{Textual Perturbation Toward ASR Robustness}
The majority of prior research involving the creation of semantic or textual perturbations has been centered around emulating ASR errors~\cite{yang2020characterizing, yang2022mitigating}. These perturbations are then applied to synthetically generated data to enhance the robustness of subsequent natural language understanding (NLU) tasks, such as speech translation~\cite{li2018improving}, entity resolution~\cite{zhou2022phonetic, yang2021voice2series}, and dialog act classification~\cite{wang2020data}. Unlike these approaches, our study is based on introducing alterations to the $N$-best output from the initial acoustic model pass. The emphasis is on producing disturbances that enhance the robustness of the second-pass rescoring model.
\label{sec:relatedwork}

\section{Methodology}
\label{sec:method}

\begin{figure*}
    \centering
    \includegraphics[scale=0.38]{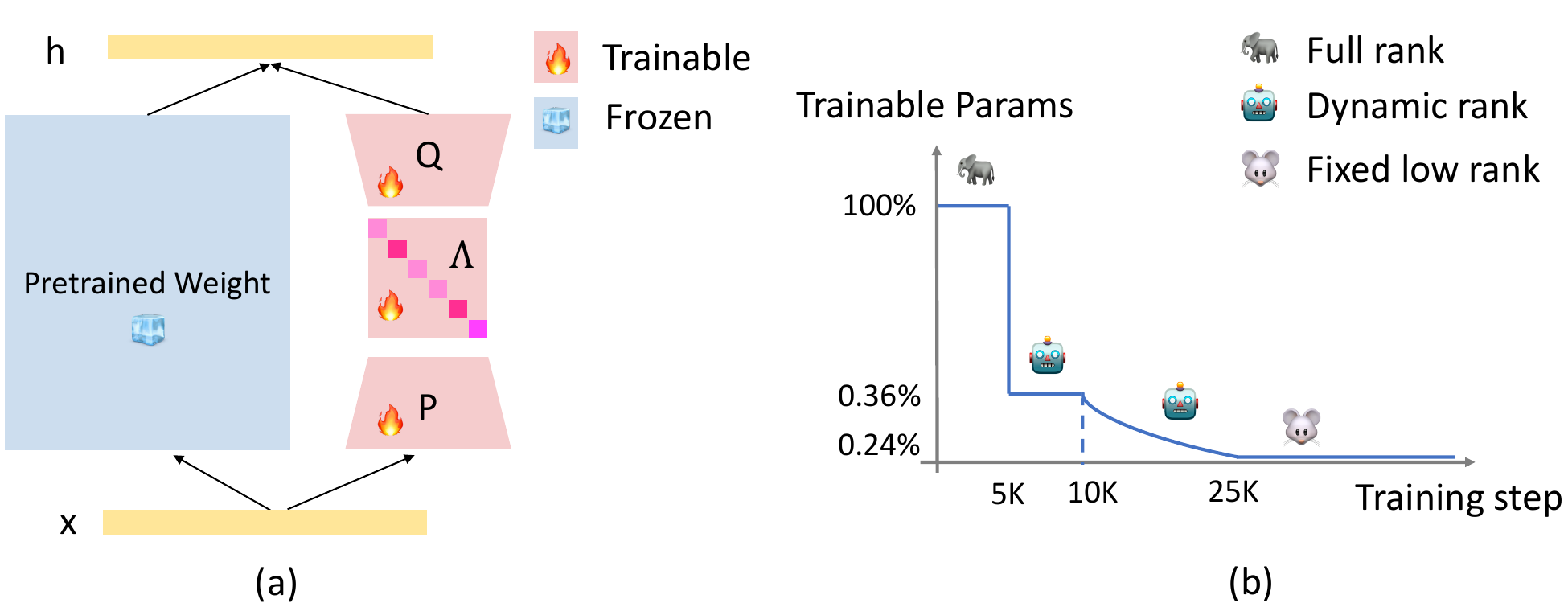}
    \caption{Two improved training strategies for LoRA-based ASR language modeling: (a) dynamic rank allocation and (b) mixed-rank training. For mixed-rank training: full rank training is marked by an elephant icon, dynamic rank allocation is marked by robots, and the very low-rank fine-tuning is marked by a mouse. }
    \label{fig:train-strategy}
\end{figure*}

\subsection{Low rank adaptation strategies}
\subsubsection{LoRA: Low-rank adaptation}
LoRA~\cite{hu2021lora} decomposes the incremental update of the pretrained weights into two matrices $W_A$ and $W_B$, where $W_0 \in R^{d1\times d2}$, $W_A \in R^{r\times d2}$ and $W_B \in R^{d1\times r}$. Given a hidden representation $h = W_0 x$, the low-rank adapted representation becomes $h = W_0 x + \Delta x = W_0 x + W_B W_A x$.  $W_A$ is initialized to a Gaussian distribution, and $W_B$ is initialized with zero to ensure $\Delta=0$ at the start of training.

\subsubsection{Strategy $1$ ($\mathcal{S}_1$) for dynamic rank allocation}
Dynamic search of neural network architectures, such as sparse reparameterization~\cite{dai2019nest,mocanu2018scalable,mostafa2019parameter}, has been investigated in previous work for parameter-efficient training and better generalization performance. Among them, dynamic rank allocation~\cite{zhang2023adaptive} is a method to adaptively search for optimal network structure during fine-tuning. Dynamic rank allocation  models the incremental update of the pretrained weights into three matrices $P$, $\Lambda$, and $Q$, where $P \in R^{d1\times r}$, $\Lambda \in R^{r\times r}$ and $Q \in R^{r\times d2}$. Denote the total rank budget as $B$, which is the product of the target rank, the number of adapted matrices in each layer, and the total number of layers. For example, if inserting low-rank matrices into every pretrained weight (e.g., $W_q$, $W_k$, $W_v$, $W_o$, $W_{f1}$, $W_{f2}$) in a 12-layer rescoring BERT with a target rank of 8 and a rank budget $B=8 \times 6 \times 12=576$.  For the $k$-th adapted weight matrix, the singular value in the $i$th dimension is denoted by $\Lambda_{k}^i$, and the singular vectors in the $i$th dimension are denoted as $P_{k}^i$ and $Q_{k}^i$ respectively. For each triplet $(P_{k}^i, \Lambda_{k}^i,Q_{k}^i)$, an importance score $s_k^i = I(P_{k}^i, \Lambda_{k}^i,Q_{k}^i)$ is computed, following the sensitivity score definition $I = |w \frac{\partial L}{\partial w}|$, which measures how much the loss will change by pruning the weight $w$. Then all importance scores $s_k^i$ will be sorted, and the rank of the triplet is compared with the current budget $B$: if the rank is larger than $B$, the value of $\Lambda_{k}^i$ is pruned to zero; otherwise the $\Lambda_{k}^i$ value is kept.

\subsubsection{Strategy $2$ ($\mathcal{S}_2$) for high rank warm-up}
High-rank warm-up before low-rank adaptation has been shown to be an effective training strategy for reducing the performance gap between LoRA and full fine-tuning in the scenario of pretraining language models~\cite{lialin2023stack}. Follow~\cite{lialin2023stack}, we  unfreeze all trainable parameters in the pretrained language model for $5000$ training steps and then start the standard low-rank fine-tuning.

\subsubsection{Strategy $3$ ($\mathcal{S}_3$) for mixed-rank staging}
Combining strategies 1 and 2, we propose a new training scheduler that controls the rank of incremental weight matrices on the fly depending on the training step, as shown by Equation ~\ref{schedule}. We split the training into four stages and use $t^w$ (full rank warm-up), $t^i$ (rank allocation initialization), $t^f$ (rank allocation finalization), and $T$ (normal fine-tuning training) to denote the final training step of each stage. Similarly,  $r^f$,  $r^i$,   $r^T$  denote the full-rank, dynamic allocation initialized rank, and target low-rank, respectively. We follow \cite{zhang2023adaptive} in choosing $r^i$ to be 1.5 times larger than $r^T$.

\begin{equation}
\label{schedule}
     r(t)=\begin{cases}
               r^f   & 0\leq t < t^w\\
               r^i   & t^w\leq t < t^i\\
     r^T + (r^i-r^T)(1-\frac{t-t^i-t^f}{T-t^i-t^f})^3   & t^i\leq t < T-t^f\\
              r^T   & t \geq T-t^f \\
            \end{cases}
\end{equation}
 
\subsection{N-best Perturbation based Rescoring Robustness}

\subsubsection{N-best perturbation}
Our goal is to create noisy examples with input perturbations on the $N$-best hypotheses, such that the ranks of the hypotheses will be corrupted and the weaknesses of the rescoring model can be exposed. 

Previously, adding ``naturally occurring noise'', such as spelling errors on inputs of machine translation, has been shown to be effective at improving the robustness of the transformer-based machine translation model~\cite{karpukhin2019training, niu2020evaluating}. Similarly, we aim to identify the errors occurring naturally in $N$-best input. To this end, we analyze the internal ``low-resource music domain'' data and categorize the most frequent errors that are related to 1) white space insertion, such as ``maybe $\rightarrow$ may be'', 2) token/character replacement, such as ``Wang Jian $\rightarrow$ Wang Jiao'', and 3) homophone replacement, such as ``you're $\rightarrow$ your''. To summarize, all three types involve replacement by phonetically similar phrases.

\begin{figure}
    \centering
    \includegraphics[scale=0.33]{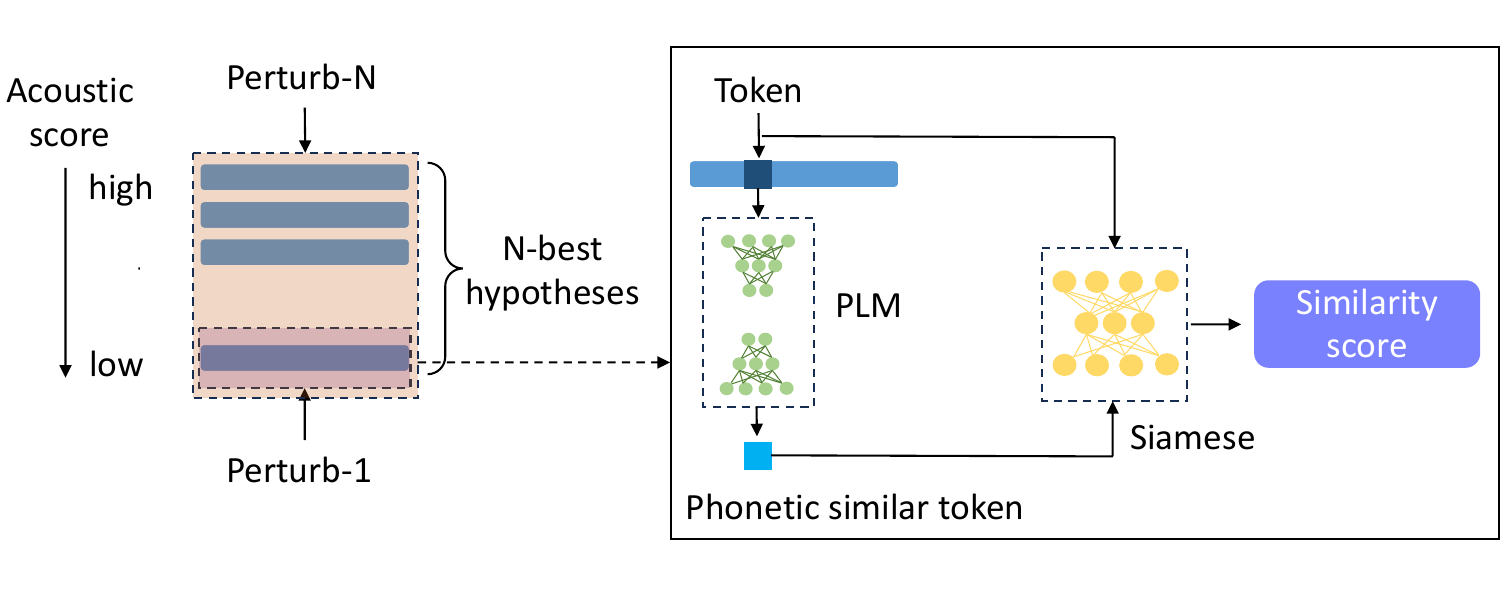}
    \caption{Proposed N-best evaluation for robust ASR Rescoring.}
    \label{fig:n-best_perturb}
\end{figure}

\subsubsection{Phonetics-based perturbation generation}

Drawing inspiration from ~\cite{eger2020hero}, we generate such phonetically similar perturbations for each word following two steps. In the first step, the phonetic word representation of one token $w$ is generated by a Seq2Seq model. Then, the second Seq2Seq model converts the phonetic word representation to a sound-alike word $\hat{w}$. Both Seq2Seq models are trained on the Combilex dataset~\cite{richmond2010generating}. In the second step, a Siamese network of InferSent~\cite{conneau2017supervised} is used to detect whether $w$ and $\hat{w}$ are phonetically similar.
The InferSent network is trained to render two word representations close if they are pronounced alike.

We apply the phonetics-based perturbations to a single hypothesis, denoted by \textit{perturb-$1$}, or all hypotheses, denoted by \textit{perturb-$N$}. In the scenario of single-hypothesis perturbation, we consider that one naturally errorful hypothesis is mistakenly generated by the first-pass acoustic model, so we apply perturbation to the hypothesis with the lowest acoustic score obtained from the first-pass acoustic model. In the scenario of all-hypothesis perturbation, we consider the rescoring model independent of the acoustic model, so we apply perturbation to each hypothesis in the $N$-best list.
In both perturbation scenarios, each token is replaced with a probability of $0.5$.

\subsubsection{Robustness metric}
We now define the N-best Perturbation-based Rescoring Robustness (NPRR) evaluation metric. Denote the rescoring model by $f$, the N-best input by $X$, the perturbed N-best input by $X'$, and the reference transcription as $Y$. The NPRR metric takes two types of degradation into consideration: 1) the absolute degradation $\Delta$ WER relative to the oracle word error rate, which is the upper bound for the WER a rescoring model could achieve; and 2) the relative degradation compared to the clean N-best input.

\begin{equation}
    \Delta \text{WER}(f,X',Y) = \text{WER}(f, X', Y) - \text{OracleWER}(f, X', Y)
\end{equation}

\begin{equation}
    \text{NPRR}(f,X,X',Y) = \frac{\Delta \text{WER}(f,X',Y)-\Delta \text{WER}(f,X,Y) }{\Delta \text{WER}(f,X,Y)}
\end{equation}

\section{Experiments}
\label{sec:experiment}
\begin{table}[!h]
\caption{Relative WER improvement of full fine-tuning (FT), vanilla LoRA, and Strategies 1, 2, and 3 on internal messaging data, where users are \textbf{not} identifiable with an absolute \textit{internal baseline} WER $<$9\%.} 
\label{tab:messaging_wer}
\adjustbox{width=0.48\textwidth}{
\centering
\begin{tabular}{l|c|c}
\toprule

Method & Trainable params& WER Relative ($\uparrow$)\\\hline
Rescore BERT (non adapted)  & 170M &\textit{internal baseline}  \\\midrule
Fine-Tuning (FT)  &100\% &3.30\%   \\\midrule
vanilla LoRA&0.24\% &7.45\% \\
w/ $\mathcal{S}_1$: dynamic rank &0.36\% $\rightarrow$ 0.24\% &\textbf{11.12\% }\\
w/ $\mathcal{S}_2$: warm-up &100\% $\rightarrow$ 0.24\% &8.47\% \\
w/ $\mathcal{S}_3$: staging ($\mathcal{S}_1$+$\mathcal{S}_2$) &100\% $\rightarrow$ 0.36\% $\rightarrow$ 0.24\% &10.57\% \\

\bottomrule
\end{tabular}}
\end{table}
\subsection{Implementations}
For all adaptation experiments, the implementation is based on the publicly  released \textit{Huggingface PEFT} \cite{peft} code base. To evaluate the performance of low-rank adaptation and the advanced three strategies, we fine-tune a 170M parameter rescoring BERT on a de-identified in-house messaging dataset (240K utterances). Besides, for a fair comparison with previous work~\cite{yu2023low}, we also fine-tune a publicly released BERT-base-cased model on the public Librispeech dataset. In the Vanilla LoRA baseline experiment, we use cross-validation to choose the hyper-parameters and fix the LoRA rank to $8$, LoRA dropout rate to $0.1$, and LoRA $\alpha=32$. For Strategy 1 (\textit{dynamic rank allocation}), we follow \cite{zhang2023adaptive} in settting the initial rank to $12$, the target rank to $8$, the training steps using initial ranks to $5000$, the starting training step using target rank to $25000$, and target modules to all weight matrices $W_q$, $W_k$, $W_v$, $W_o$, $W_{f1}$, and $W_{f2}$.
For Strategy 2 (\textit{high rank warm-up}) and Strategy 3 (\textit{mixed-rank training}), we first unfreeze all parameters in the pretrained models for 5000 training steps, then start low-rank training or dynamic rank allocation training.  

\begin{table}[!h]
\caption{Absolute WER on the two standard test sets of the LibriSpeech corpus~\cite{panayotov2015librispeech}, decoded by Whisper-tiny. All low-rank adaptation results are obtained by tuning the coefficient~\cite{xu2022rescorebert} of second-pass rescoring scores. The 170M BERT base-cased model is retrieved from official public release~\cite{devlin2019bert} for reproducible evaluation.} 
\label{tab:librispeech_wer}

\adjustbox{width=0.49\textwidth}{
\centering
\begin{tabular}{l|c|c|c}
\toprule
    Model and method  &\% Trainable params. & test-clean & test-other \\\midrule
    BERT$_\text{base-cased}$& non-adapted &6.17 & 13.81 \\\hline
    w/ FT& 100\% &4.87 &12.47  \\ \hline
    vanilla LoRA&0.27\% &4.78 &12.21 \\
    w/ $\mathcal{S}_1$: dynamic rank&0.4\% $\rightarrow$ 0.27\% &4.71  &\textbf{12.11}  \\
    w/ $\mathcal{S}_2$: warm-up & 100\%$\rightarrow$  0.27\%&4.76  &12.15  \\
    w/ $\mathcal{S}_3$: staging ($\mathcal{S}_1$+$\mathcal{S}_2$)& 100\%$\rightarrow$ 0.4\%$\rightarrow$  0.27\% &\textbf{4.69}  & 12.17 \\

    \bottomrule
\end{tabular}}
\end{table}
\begin{table*}[ht!]
\caption{Robustness evaluation of ASR-LM under $N$-best perturbations tested on the LibriSpeech baselines reported in Table~\ref{tab:librispeech_wer}. } 
\label{tab:robustness}

\adjustbox{scale=0.75}{
\centering
\begin{tabular}{l|c|c|c|c|c|c|c|c|c|c}
\toprule
Pretrained BERTs &
  Test set &
  WER $\downarrow $ &
  OracleWER $\downarrow $ &
  $\Delta$ WER $\downarrow $ &
  NPRR[\%] $\downarrow $ &
  Test set &
  WER$\downarrow $ &
  OracleWER $\downarrow $ &
  $\Delta$ WER $\downarrow $ &
  NPRR[\%] $\downarrow $ \\ \hline
Fine-tune based                     & test-clean     & 4.87 & 3.59 & 1.28 & -      & test-other     & 12.47 & 9.97  & 2.50 & -                            \\ 
                              & w/ perturb-$1$ & 5.10 & 3.60 & 1.50 & \cellcolor[HTML]{EFEFEF}17.18  & w/ perturb-$1$ & 12.82 & 10.09 & 2.73 & \cellcolor[HTML]{EFEFEF}9.20 \\ 
 &
  w/ perturb-$N$ &
  5.37 &
  3.56 &
  1.81 &
  41.40 &
  w/ perturb-$N$ &
  13.07 &
  9.92 &
  3.15 &
  26.00 \\ \hline
vanilla LoRA                          & test-clean     & 4.78 & 3.43 & 1.35 & -      & test-other     & 12.21 & 9.77  & 2.44 & -                            \\ 
                              & w/ perturb-$1$ & 5.03 & 3.44 & 1.59 & \cellcolor[HTML]{9AFF99}\textbf{17.73}  & w/ perturb-$1$ & 12.58 & 9.89  & 2.69 & \cellcolor[HTML]{9AFF99}\textbf{10.44}                        \\ 
                              & w/ perturb-$N$ & 6.27 & 3.40 & 2.87 & 112.76 & w/ perturb-$N$ & 14.77 & 9.72  & 5.05 & 106.82                       \\ \hline
LoRA based on $\mathcal{S}_1$ & test-clean     & 4.71 & 3.44 & 1.27 & -      & test-other     & 12.11 & 9.67  & 2.44 & -                            \\ 
                              & w/ perturb-$1$ & 5.66 & 3.45 & 2.21 & 74.24  & w/ perturb-$1$ & 13.65 & 9.79  & 3.86 & 57.37                        \\ 
                              & w/ perturb-$N$ & 5.39 & 3.41 & 1.98 & \cellcolor[HTML]{9AFF99}\textbf{56.06 } & w/ perturb-$N$ & 13.35 & 9.63  & 3.73 & 52.98                        \\ \hline
LoRA based on $\mathcal{S}_3$ & test-clean     & 4.69 & 3.44 & 1.25 & -      & test-other     & 12.17 & 9.72  & 2.45 & -                            \\ 
                              & w/ perturb-$1$ & 5.48 & 3.45 & 2.02 & 62.30  & w/ perturb-$1$ & 13.10 & 9.84  & 3.26 & 33.06                        \\ 
                              & w/ perturb-$N$ & 5.66 & 3.41 & 2.24 & 80.00  & w/ perturb-$N$ & 13.15 & 9.67  & 3.47 & \cellcolor[HTML]{9AFF99}\textbf{41.83}                        \\ 
    
    \bottomrule
\end{tabular}}
\end{table*}

\subsection{Performance evaluation}

\textbf{Performance on messaging data}: The relative word error rate improvement on the test set of messaging data is shown in Table~\ref{tab:messaging_wer}. All low-rank training methods outperform full fine-tuning on this specific dataset. In contrast to vanilla LoRA, the three advanced strategies show consistent improvements, with dynamic rank allocation leading with 3.67\% relative gain. Notably, when compared to  Strategy 2 \textit{high-rank warm-up} technique, Strategy 1 \textit{dynamic rank allocation} proves to be a more effective approach for the fine-tuning process of the pretrained rescoring BERT model.


\textbf{Performance on Librispeech:} The relative improvement in word error rate on the test sets of Librispeech is shown in Table~\ref{tab:librispeech_wer}. Consistent with the results on the messaging data, the \textit{dynamic rank allocation} technique stands out by delivering the most substantial reduction in word error rate (i.e., 2\% improvement over FT) across both the test-Clean and test-Other subsets. Remarkably, when \textit{dynamic rank allocation} is paired with \textit{high-rank warm-up}, there is an incremental, yet notable, improvement observed in fine-tuning the \textit{non-rescoring} BERT-base-cased model. Our investigation, focusing on both the rescoring BERT and the BERT-base-cased model, underscores that \textit{high-rank warm-up} is primarily advantageous in the context of pretraining the rescoring model, rather than being impactful in the incremental learning phase of the rescoring model.

\textbf{A case study on layer-wise performance:} We conduct an in-depth layer-wise low-rank adaptation experiment to examine each layer's importance with regard to rescoring performance. The word error rate results of applying LoRA to each layer in BERT-base-cased is presented in Table~\ref{tab:librispeech_layer_wer}. The intermediate four layers perform considerably better than the first and last four layers. Among the intermediate four layers, layer $5$ and layer $6$ achieve the best word error rate on the Test-Clean and the Test-Other set, respectively. Given that prior research has indicated that intermediate layers of BERT hold and process syntactic information~\cite{jawahar2019does,merchant2020happens}, we can infer that acquiring syntactic features (such as subject-verb agreement) from the fine-tuning data plays a crucial role in the success of the rescoring task. Interestingly, when we scale up to the 1B model, we observe the best results for the first two layers ($0$, $1$), second best for the middle layers ($5$, $6$), and the worst results for the last two layers ($10$, $11$). We also observe higher WER for the 1B model, which might be attributed to overfitting.
\begin{table}[!h]
\caption{170M BERT single layer-wise performance.} 
\label{tab:librispeech_layer_wer}
\centering
\adjustbox{width=0.38\textwidth}{
\centering
\begin{tabular}{l|c|c}
\toprule
    Model / Layer  & Test-Clean & Test-Other \\\midrule
    Fine-tuning BERT$_\text{base}$ &6.17 & 13.81 \\\hline
    vanilla LoRA all layers &4.78 &12.21  \\ \hline
    LoRA-adapted layer 0 &5.39 &13.05  \\ 
    LoRA-adapted layer 1 &5.33 &12.87 \\
    LoRA-adapted layer 2 &5.28  &12.88  \\
    LoRA-adapted layer 3 &5.30  &12.82  \\
    LoRA-adapted layer 4 &5.25  &12.71  \\ 
    LoRA-adapted layer 5 &5.28  &\textbf{12.66}  \\ 
    LoRA-adapted layer 6 &\textbf{5.21}  &12.82  \\ 
    LoRA-adapted layer 7 &5.24  &12.84  \\ 
    LoRA-adapted layer 8 &5.30  & 12.78 \\ 
    LoRA-adapted layer 9 &5.33  & 12.93 \\ 
    LoRA-adapted layer 10 & 5.29 &12.99  \\ 
    LoRA-adapted layer 11 &5.31  &12.95  \\ 
    \bottomrule
\end{tabular}
}
\end{table}



\subsection{Robustness evaluation}
The evaluation of robustness under $N$-best input perturbations is presented in Table~\ref{tab:robustness}, allowing us some key takeaways. First, applying phonetic similarity-based token replacement to all $N$-best hypotheses improves the oracle WER, e.g., $3.59 \rightarrow 3.56, 9.97 \rightarrow 9.92$. This shift signifies an improvement in the quality of input within the $N$-best hypotheses. Ideally, this points to a robust rescoring model yielding reduced WER post-training. Nonetheless, it is striking that all fine-tuned models exhibit varying degrees of vulnerability in adapting to such ``positive'' perturbations, displaying distinct levels of degradation. Second, the fully fine-tuned model achieves the lowest \textbf{NPRR} score and thus is the most robust among all models under both perturbing strategies. Interestingly, the behavior of the vanilla low-rank adapted model diverges markedly under the two perturbation strategies. When just a single hypothesis is perturbed, it mirrors the response of the fully fine-tuned model. Conversely, under perturb-$N$, its performance is completely corrupted, particularly in cases where the quality of $N$-best hypotheses is promoted. This discrepancy hints at the sensitivity of the LoRA model to the $N$-best input derived from the first-pass acoustic model. Consequently, any alteration to the acoustic model could potentially result in an unsatisfactory performance for the LoRA rescoring model. Finally, we observe adding \textit{dynamic rank allocation} can degrade more in \textbf{NPRR} than with vanilla LoRA. 




\section{Conclusion}
\label{sec:conclusion}
We have conducted a comprehensive evaluation of low-rank adaptation fine-tuning and its advanced variants. The experimental results show that \textit{dynamic rank allocation} yields a further enhancement of 3.67\% beyond the performance of LoRA. However, when considering the case study involving the adoption of a low-rank adapted rescoring BERT to assess adversarial robustness within the context of $N$-best input perturbations, this particular strategy intensifies the degradation of the robustness score. In the future, we intend to delve into the generation of input perturbations based on generative language models and harnessing synthetic data augmentation as a robust training mechanism for the rescoring model.

\subsection*{Acknowledgment} The authors thank Qi Luo, Aditya Gourav, Yile Gu, Yi-Chieh Liu, Shanili Ghosh, I-Fan Chen, Mat Hans, Grant Strimel, and Bjorn Hoffmeister for their discussions and valuable feedback.


\clearpage
\bibliographystyle{IEEEbib}
\bibliography{strings,refs}

\end{document}